\def\year{2022}\relax
\documentclass[letterpaper]{article} 
\usepackage{aaai22}  
\usepackage{times}  
\usepackage{helvet}  
\usepackage{courier}  
\usepackage[hyphens]{url}  
\usepackage{graphicx} 
\usepackage{natbib}  
\usepackage{caption} 
\DeclareCaptionStyle{ruled}{labelfont=normalfont,labelsep=colon,strut=off} 
\usepackage[switch]{lineno} 
\usepackage{amssymb}
\usepackage{amsmath}
\usepackage{float}
\usepackage{tabularx}
\usepackage{caption}
\usepackage[ruled,vlined]{algorithm2e}
\usepackage{amsmath}
\usepackage{booktabs}
\usepackage{footnote}
\usepackage{tablefootnote}
\usepackage{subfigure}
\usepackage{hyperref}

\title{Taming Repetition in Dialogue Generation}
\author{Yadong Xi, Jiashu Pu, Xiaoxi Mao\\
  {\normalsize Fuxi AI Lab, NetEase Inc., Hangzhou, China} \\
  \texttt{\normalsize \{xiyadong, pujiashu, maoxiaoxi\}@corp.netease.com}}

\usepackage{bibentry}

\begin{document}

\maketitle

\begin{abstract}
The wave of pre-training language models has been continuously improving the quality of the machine-generated conversations, however, some of the generated responses still suffer from excessive repetition, sometimes repeating words from utterance, sometimes repeating words within self-generated responses, or both. Inappropriate repetition of words can significantly degrade the quality of the generated texts. Penalized sampling is one popular solution, reducing the sampling probability of existing words during inference, however, it is highly vulnerable to the inappropriate setting of the static weight. Setting it too high can yield strange and unrealistic sentences while setting it too low makes the task of suppressing repetition trivial. To remedy the shortcomings of the above methods, we design a context-aware classifier to explicitly decide when to allow repetition and when to employ penalized sampling. Such a classifier can be easily integrated with existing decoding methods, reducing repetitions where appropriate while preserving the diversity of the text. Experimental results demonstrate that our method can generate higher quality and more authentic dialogues.
\end{abstract}

\section{Introduction}

\begin{figure*}[tbh]
    \centering
\includegraphics[width=0.75\textwidth]{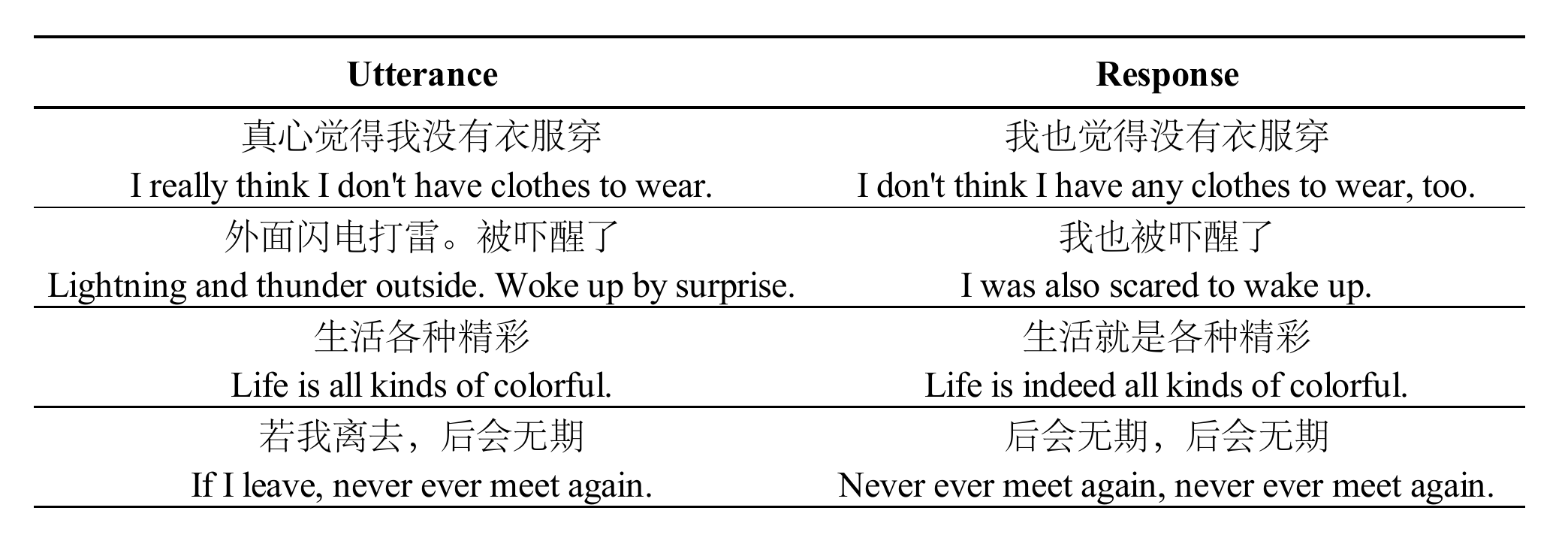}
    \caption{Examples of dialogue responses with excessive repetitions.}
    \label{fig:case}
\end{figure*}

The dialogue generation task is taking a big step forward with the help of the pre-training paradigm~\citep{transfertransfo,meena,blenderbot}. However, excessive repetition is still common in generated responses due to the limitation of the sampling methods~\citep{degeneration}. Figure~\ref{fig:case} shows some typical examples of excessive repetition from our dialogue generation model, although the responses are suitable for the utterance, they are not informative and interesting enough. When having conversations with such repeat-prone chatbots, users are likely to lose interest soon.

Although~\citet{massive} proves the repetition problem disappears when randomly sampling from an untruncated distribution, the quality of the text cannot be guaranteed. Instead, some researchers prevent words from appearing twice by enforcing a static penalty to those words that have already appeared in the previous dialogue context~\citep{opennmt,importance,what,ctrl}. However, searching proper hyperparameters for these methods is non-trivial. In practice, we find that setting an inappropriate static penalty can cause problems in two aspects. First, setting the static penalty too high may unexpectedly increase the words that are semantically related but not context-aware, thus producing unrealistic responses. For instance, for utterance \textbf{Person A:}``I wanna \emph{hot} dog.'', model may generate \textbf{Person B:}``There is no \emph{cold} dog left'' because the \emph{hot} and \emph{cold} are antonym.  In addition, the redistribution of sampling probability of existing words makes the model more inclined to sample high-frequency words. For example, dull responses such as \textit{I don't know or I'm OK} are more likely to be generated. Secondly, setting the static penalty too low may arise the repetition feedback loops~\citep{degeneration}, which makes the process of repetition control trivial. The above evidence shows simple rule-based repetition control methods are not powerful enough to finely manage the repetition in dialogue generation.

Popular decoding strategies, such as top-k or top-p sampling~\citep{holtzman2019curious}, are proposed to generate more fluent texts. However, they output a truncated (distorted) distribution during inference, which is inconsistent with training. In this work, we aim to guide the decoding process with a non-distorted signal, a signal that is generally consistent between training and inference. Specifically, we train a classifier conditioned on the previous context to explicitly control repetitions. The (not)-allow-to-repeat signal is essentially the categorical output of the classifier. The repetition can occur in two situations, either a word to be sampled is already in the utterance (partner repetition) or the generated response (self-repetition). We choose to build two different classifiers to handle both situations separately. Furthermore, we analyze the repetition phenomenon in the gold data and find repetition statistics is weakly related to the utterance context. The independence between the repetition statistics and the context motivates us to design two embedding layers to incorporate repetition statistics as additional inputs. In essence, our method is equivalent to augmenting the penalized sampling with context knowledge, which alleviates the risk of generating semantic implausible phrases or dull responses. We summarize our contributions as follows:
\begin{itemize}
  \item We shed light on the side-effects of the penalized sampling in dialogue generation.
  \item We introduce a context-aware classifier to control the repetition, assisting the dialogue generation model in decoding better quality texts. The repetition-control classifier is fully compatible with mainstream decoding strategies, it can be easily integrated and is proved effective in reducing repetition without losing text diversity.
  \item We construct a Chinese dialogue dataset where human and automatic evaluations are conducted. Compared with other baselines, our framework generates dialogues of better quality.\footnote{The dialogue dataset and pre-trained language model will be open sourced upon publication.}
\end{itemize}

\begin{figure*}[thb]
  \centering
  \includegraphics[scale=0.43]{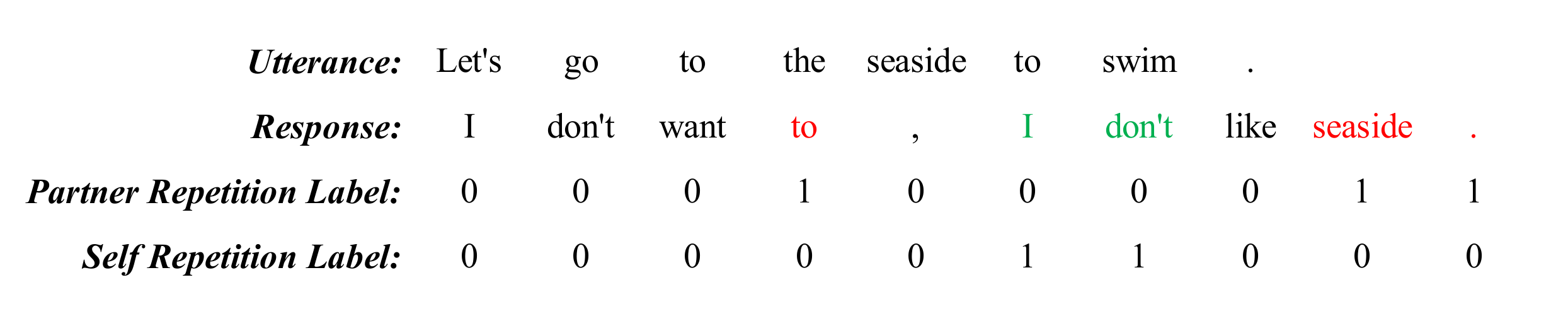}
\caption{\small \textbf{Illustration of the repetition labels.} The words in red and green mean they have already appeared in utterance and previous decoded responses respectively. Labels for Partner and self-repetition are bool, with label $0$ corresponding to ``no repetition found'' and label $1$ corresponding to ``the word repeats at least once''.}
\label{fig:label}
\end{figure*}

\begin{figure*}[thb]
  \centering
  \includegraphics[scale=0.6]{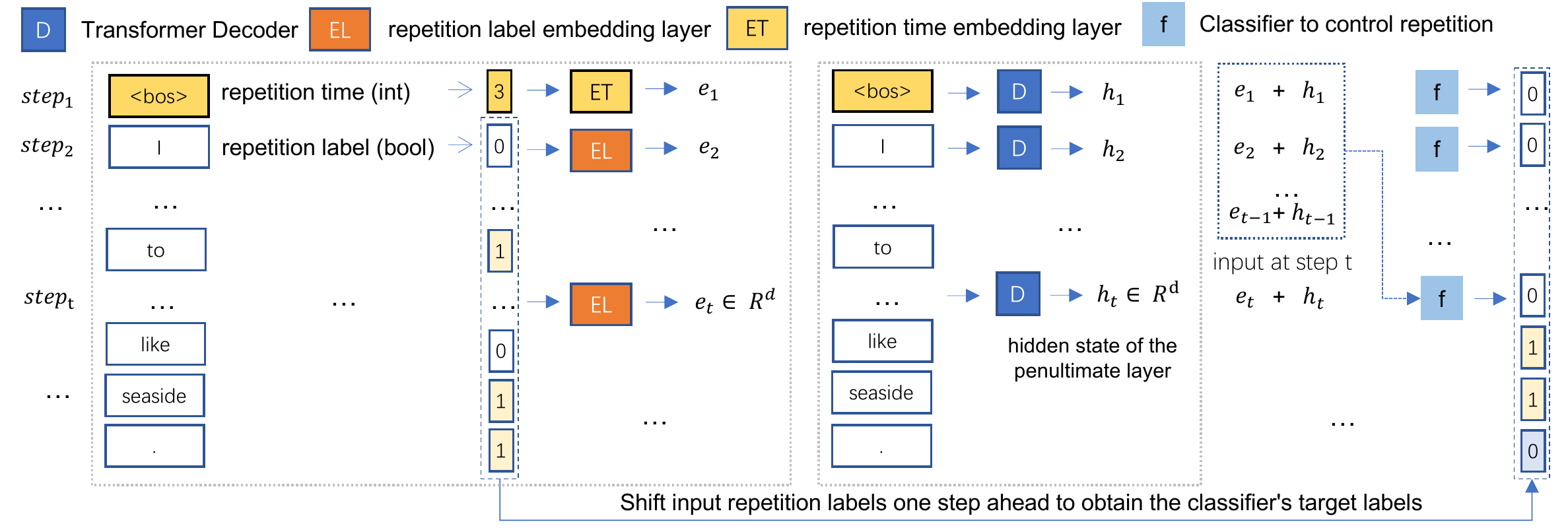}
\caption{\small \textbf{Illustration of the partner classifier's input stream.} Each token’s input is the sum of the hidden state of decoder's penultimate layer and a repetition label/time embedding. The same input format applies to the self-repetition classifier as well except that values of the repetition label and time need to be recalculated accordingly.}
\label{fig:input}
\end{figure*}

\section{Related Works}

\subsection{Dialogue Generation}
The dialogue generation models based on deep learning have made remarkable progress~\citep{seq,gan,attention}. In recent years, the performance of a range of natural language processing tasks has been promoted by the pre-training methods with a large margin~\citep{gpt,bert,gpt3}. The pre-trained language models are capable of generating responses with high quality by fine-tuning on a small dialogue dataset~\citep{transfertransfo,lost}. Meanwhile, pre-training a large-scale language model on a huge dataset with a bulk of dialogues is the state-of-the-art~\citep{dialogpt,meena,blenderbot}. Our method is implemented on the basis of the most relevant work~\citep{lost}.

\subsection{Repetition in Dialogue Generation}
Appropriate repetition in the text is a normal phenomenon while excessive repetition is an obstacle for natural language generation~\citep{degeneration}. The \textit{n-gram} block is a general recipe~\citep{opennmt} to tackle the issue. The repetition in dialogue is firstly explored in-depth by~\citep{what}. Instead of prohibiting the \textit{n-gram} repeating, they reduce the probability of the \textit{n-gram}s with a right weight. \textit{Penalized sampling} is proposed to discount the scores of the previously generated tokens \citep{ctrl} with a similar motivation to~\citep{what}. \citep{unlikely} proposed a new objective, \textit{unlikelihood training} to restrain the repeats and frequent words. 

In machine translation and summarization, the coverage mechanism is proposed to prevent excessive repetition with the assumption that repetition is from the attention over the same position in sequential decoding steps~\citep{tucoverage,micoverage,get,bottom}. The method is designed to avoid generating a \textit{n-gram} twice and more, which is not a mismatch for dialogue generation obviously. \citep{cutting} estimated the upper-bound frequency of each target vocabulary in the input and decoding procedure is controlled conditioned on the estimation. According to our preceding explanation, the repetition signal cannot be predicted by the utterance solely, thus it is impractical for dialogue generation.

\section{Approach}
\subsection{Base Model}
\label{base}
We follow a popular dialogue generation model architecture --- a seq2seq model, where the weights of encoder and decoder are shared \citep{lost}. The weights are initialized with a pre-trained GPT model \citep{gpt}. The loss function for training is formally defined below. Given the dialogue context $C$, the target response $T$, we have
\begin{equation}
Loss = L_{dec} + \lambda_{lm} \cdot L_{enc}
\label{eq:m1}
\end{equation}

\begin{equation}
L_{enc} = -\sum_i{logP(C_i|C_{1},...,C_{i-1})}
\end{equation}

\begin{equation}
L_{dec} = -\sum_i{logP(T_i|T_{1},...,T_{i-1},C)}
\end{equation}
, where $L_{enc}$ and $L_{dec}$ are the maximum likelihood loss of the input sequence and target sequence respectively. The motivation is to keep the similar optimization objective in accordance with the pre-training stage. We tune $\lambda_{lm}$ in the experiments and set to $0.2$.

\subsection{Classifier}
In the spirit of multi-task learning~\citep{multitask}, 
we design repetition-control classifiers of a simple structure. It contains one Transformer layer and a binary classification head. The Transformer layer is initialized with the last layer of a pre-trained GPT and the classification head is randomly initialized.

We empirically find the repetition phenomenon is weakly related to context, which inspired us to incorporate repetition statistics as additional inputs. We validate whether the level of repetition is related to the utterance intent by fine-tuning a Roberta \citep{roberta} to predict the ratio of overlap tokens between utterance and response, given only the utterance as input. Figure \ref{fig:copy} shows that predicting the overlap ratio is relatively difficult, which is evidenced by the difference between the true distribution and the bell-shaped distribution from the model's prediction. The result indicates repetitions are not correlated with contexts and motivates us to fuse the repetition distribution separately. We build two classifiers for two repetition scenarios, for calculating repetition in $T$ and $C$, we denote the self-repeat classifier as $f_{se}$ and the partner classifier as $f_{ut}$ respectively. At the decoding step of $t$, we can treat words' repetition information in decoded response $T_{<t}=(T_1, \dots, T_{t-1})$ as a prior signal for the classifiers. Specifically, we introduce embedding layer $EL$ and $ET$ for repetition label and repetition time respectively. We denote $RT$ as the function to calculate the number of repetitions in $T$ and $RL$ to produce a repetition label for each word $T_{i}$ (see an example in Figure~\ref{fig:label}); the binary repetition label indicates whether a word has repeated itself.

Below we formally define the input of the classifier, which is also illustrated in Figure~\ref{fig:input}. At step $t$, the embedding input of the classifier is
\begin{equation*}
(ET(RT(T_{1})), EL(RL(T_{2})), \dots, EL(RL(T_{t-1})))
\end{equation*}
while the hidden state input is
\begin{equation*}
(h_1, \dots, h_{t-1})
\end{equation*}
, where $h_i$ is the hidden state of the decoder's penultimate layer. The input of the classifier is the sum of both. At step $1$, because token is always $\langle bos \rangle$, we calculate the total number of repetitions and replace $EL$ with $ET$. Notably, because we cannot calculate the number of repetitions for future generated texts during inference, we either assign the value by sampling from a prior distribution or by a constant; the impact of different choices is illustrated in Table~\ref{tab:inference}. We also integrate the classifier loss $L_{cla}$ into the generation loss.
\begin{equation}
  Loss = L_{dec} + \lambda_{lm} \cdot L_{enc} + L_{cla}
  \label{eq:cls_loss}
\end{equation}

\subsection{Classifier Guided Decoding}
At the decoding step $t$, given the input utterance sequence $C=(c_1, \dots, c_j), c_j \in V$, the decoder, and its generated response $T_{<t}=(T_1, \dots, T_{t-1}), T_{i} \in V$ until time step $t$, the classifier $f$ outputs $y_{t} \in {0, 1}$. When $y$ equals $1$, the decoder is allowed to repeat, which means the sampling distribution of $T_{i}$ is kept intact and $T_{i}$ can be any token in vocabulary $V$; when $f_{se}$ or $f_{ut}$ outputs $0$, the decoder suppresses the sampling probability of tokens in $T_{<t}$ or $C$ respectively, specifically, the probabilities of the words appeared in the response or the utterance are set to $-inf$. When the repetition is not allowed, we further apply static penalty to sampling distribution, from which new words (tokens) are decoded. This method can be considered as a context-based \textit{1-gram} block \citep{opennmt} and can be employed to popular sampling methods, including beam search, top-k and nucleus (top-p) sampling.

\subsubsection{Determine classifiers' thresholds}
Repeating words from utterances occurs much more often than repeating words within responses, resulting in two datasets of different class ratios. To fully exploit the performance of both classifiers, we set two cut-off thresholds $h_{ut}$ and $h_{se}$ for $f_{se}$ and $f_{ut}$ separately. The setting of thresholds is based on the repetition distribution of the training data. Taking the partner classifier as an example, we first calculate the repetition ratio of a single sample on the training set, then calculate the mean of all samples' repetition ratios as the reference repetition ratio $ratio_{g}$. Afterwards, we continuously adjust the threshold $h_{ut}$ and use $f_{ut}$ to make inferences on the training set until the repetition ratio $ratio_{ut}$ of the model is close to $ratio_{g}$. The same procedure is applied to determine $h_{se}$. Alternatively, it is also possible to adjust both thresholds according to specific needs. For example, we can adjust for higher recall of the positive label, ensuring the fluency of the response.

\subsection{Classifier Enhanced Soft Penalized Sampling}
As the prediction of the classifier is not always correct, completely restricting the use of certain words in context may generate some less fluent sentences. To mitigate the impact, we try to combine the classifiers with softly penalized sampling. Below we introduce two mainstream methods of soft penalized sampling. At the step $t$ of decoding, the allow-to-repeat probability $f_{ut}(T_t)$ and $f_{se}(T_t)$ of two classifiers are available for both methods.

\citet{what} propose to modify the predicted probability of the words in the utterance $C$ and the previous decoded part $T_{<t}=T_1,...,T_{t-1}$ as:
\begin{equation}
p(C_{i})=logP(C_{i}|T_{<t},C) - w_{ut}
\label{eq:merge_2}
\end{equation}
\begin{equation}
p(T_{i})=logP(T_{i}|T_{<t},C) - w_{se} 
\end{equation}
, where $w_{ut}$ and $w_{se}$ are determined according to 
\begin{equation}
w_{ut}=\left\{
\begin{aligned}
w_{l}, \quad f_{ut}(T_t) < h_{ut} \\
w_{s}, \quad f_{ut}(T_t) \geq h_{ut}
\end{aligned}
\right.
\end{equation}

\begin{equation}
w_{se}=\left\{
\begin{aligned}
w_{l}, \quad f_{se}(T_t) < h_{se} \\
w_{s}, \quad f_{se}(T_t) \geq h_{se}
\end{aligned}
\right.
\end{equation}

\noindent Our experiments validate that setting $w_{l}$ and $w_{s}$ to 1 and 0.5 is a sound option. 

\citet{ctrl} propose another option, modifying the probability in the utterance $C$ according to 
\begin{equation}
p(C_{i})=\frac{exp(p(C_{i})/(I(C_{i}))}{\sum_{j}(exp(p(C_{j})/(I(C_{j})))}, C_{j} \in C
\label{eq:kes1}
\end{equation}
\begin{equation}
I(C_{i})=\left\{
  \begin{aligned}
  \theta_{l}, \ f_{ut}(T_t) < h_{ut} \; and \; RL(C_{i}) = 1  \\
  \theta_{s}, \ f_{ut}(T_t) \geq h_{ut} \; and \; RL(C_{i}) = 1 \\
  \quad \quad \quad 1, \quad \quad \quad         RL(C_{i}) = 0\;
  \end{aligned}
  \right.
\label{eq:kes2}
\end{equation}
, where $RL$ is denoted as the function to predict whether $C_{i}$ has repeated itself in context $C$ and the label $1/0$ indicates $C_{i}$ has been repeated or is unique. Operations in Equation~\ref{eq:kes1} and Equation~\ref{eq:kes2} are also employed to the decoded part $T_{<t}$ by replacing $C$ with $T_{<t}$. Parameters $\theta_{l}$ and $\theta_{s}$ are set to 1.25 and 1.15 in all the experiments. Notably, if there are overlap tokens between $C$ and $T_{<t}$ and both classifiers predict $0$, the probability penalizing only needs to be conducted once.

\begin{figure}[tb]
 \centering
 \subfigure[Predict repetition time given the context of utterance and response.]{\includegraphics[width=0.4\textwidth]{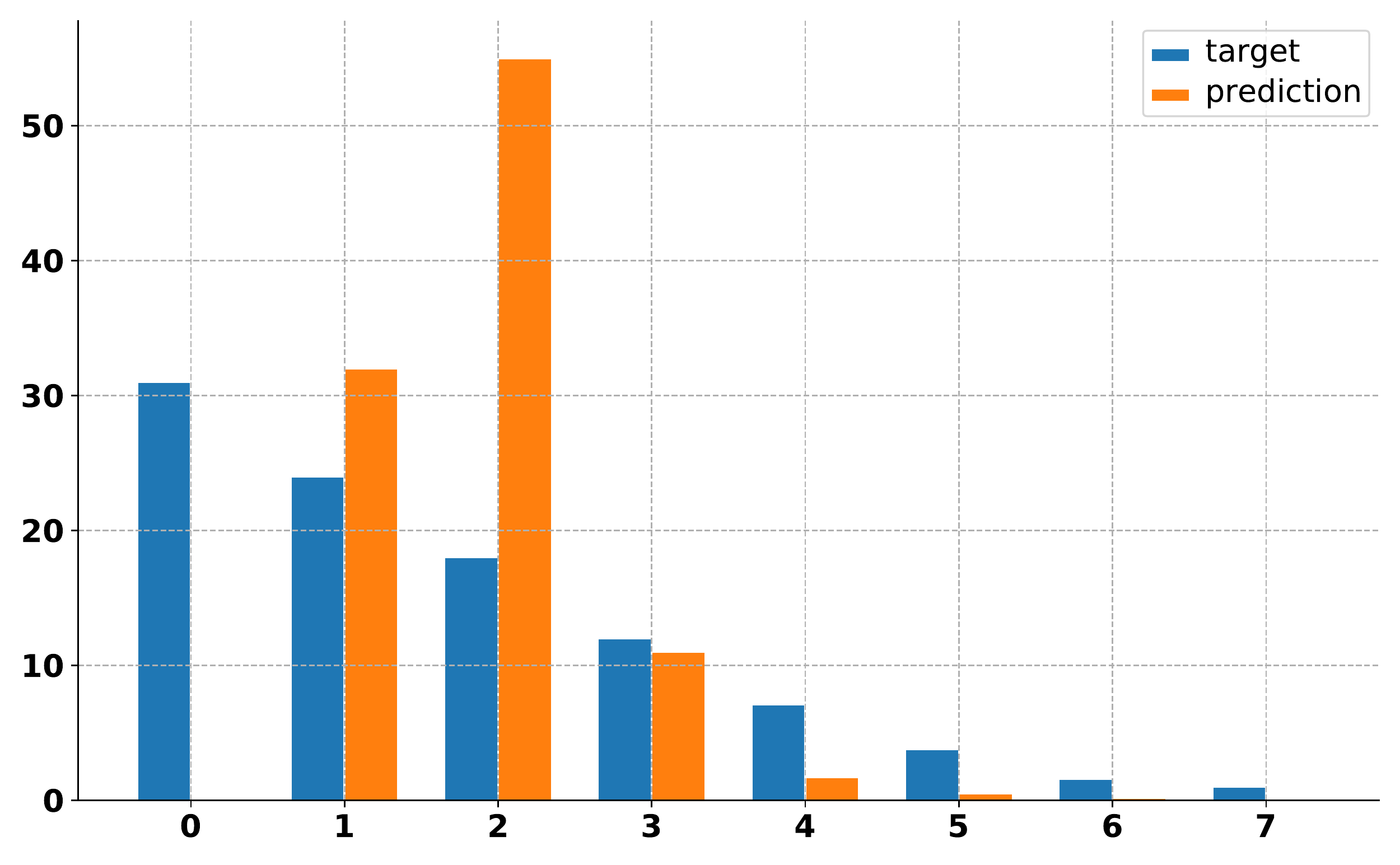}\label{fig:copy}}
 \subfigure[Ratio of the adjacent overlap score]{\includegraphics[width=0.4\textwidth]{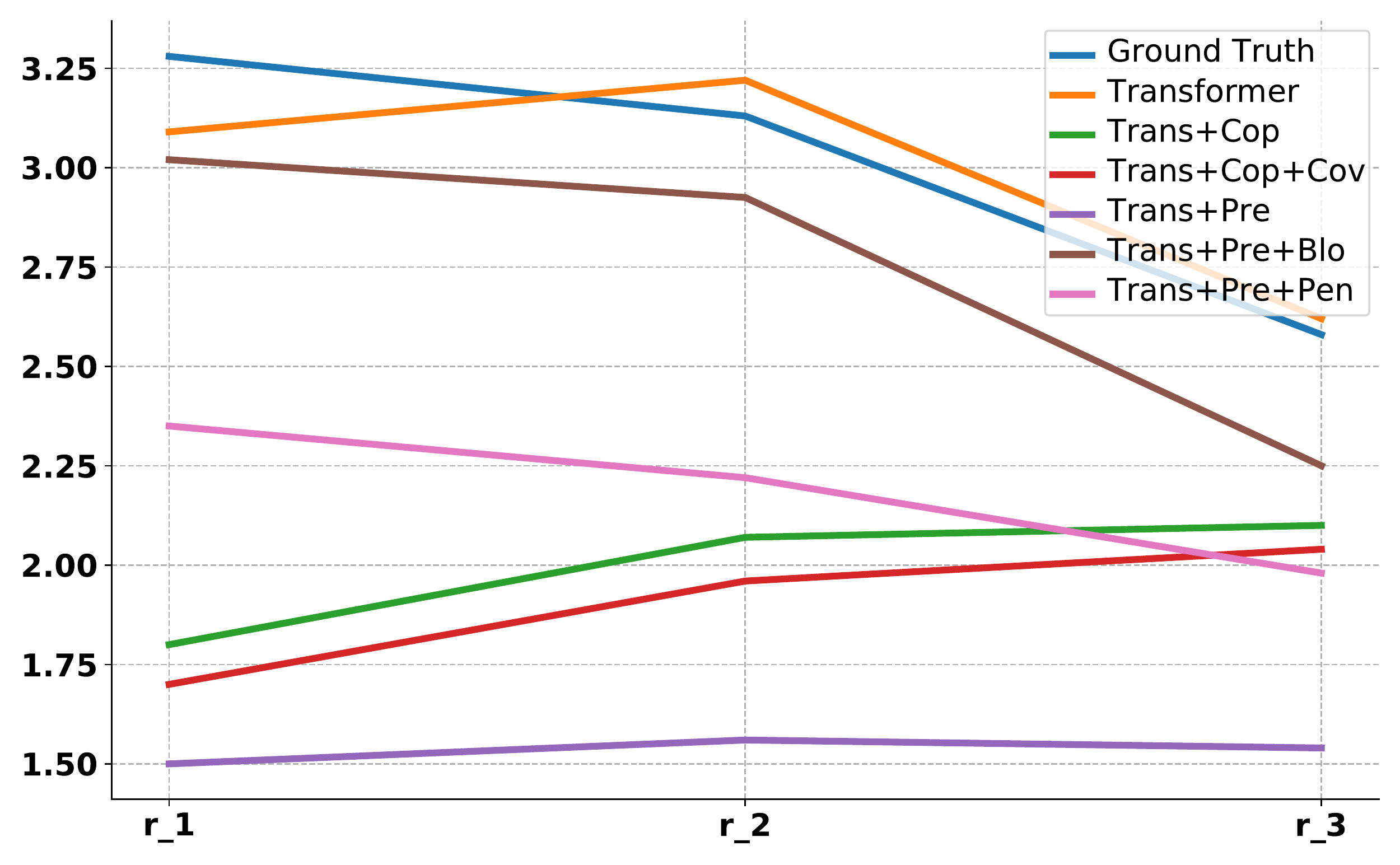}\label{fig:ratio}}
 \caption{\textbf{(a)} the x-axis presents the number of overlap tokens between the utterance and response and the y-axix is the proportion (percent) of the samples with certain repetition time. \textbf{(b)} $r_{i}$ is the ratio of the adjacent overlap score, such as $r_1=$\;{\rm \textit{ov-1}}$/${\rm \textit{ov-2}}.}
\end{figure}


\section{Experiment}

\subsection{Data and Model}
We evaluate our method and other baselines on a Chinese dialogue dataset, which is crawled from \emph{Sina Weibo}\footnote{\url{https://en.wikipedia.org/wiki/Sina_Weibo}}. 
After a rigorous manual filtering step, we obtain 520000 utterance-response pairs in total, where we randomly sample 20k pairs and 18k pairs for validation and testing. Regarding training details, the parameters of text generation models are first initialized with those of a base pre-trained GPT, then the text generation models are trained on the Sina Weibo dialogue dataset. Similar to the OpenAi-GPT~\citep{gpt}, the base GPT is pre-trained on a dataset collected from a website of Chinese novel\footnote{\url{http://www.56wen.com}}, which contains books with about 500 million tokens. The vocabulary of the base GPT only covers words that have appeared 4 times and more; it is pre-trained 70 epochs to ensure convergence.
\subsection{Baselines}
We choose several baselines to compare with our methods:

\begin{list}{$\circ$}{}
  \itemsep0em 
  \item \textbf{Transformer} \citep{attention}. Transformer treats dialogue generation as a standard seq2seq task. In experiments, we only report the results of non-pre-trained Transformers.
  \item \textbf{Trans+Cop} \citep{opennmt}. We enhance the Transformer with copy mechanism to validate its effect on dialogue generation.
  \item \textbf{Trans+Cop+Cov} \citep{bottom}. The coverage mechanism is verified effective for avoiding repetition in machine translation and abstract summarization.
  \item \textbf{Trans+Pre} \citep{lost}. A Transformer based seq2seq dialogue generation model initialized by a pre-trained GPT.
  \item \textbf{Trans+Pre+Blo} \citep{what}. Combine the \textbf{Trans+Pre} and the soft \textit{n-gram} block trick \citep{what}.
  \item \textbf{Trans+Pre+Pen} \citep{ctrl}. Combine the \textbf{Trans+Pre} and the probability penalty trick in \citep{ctrl}.
\end{list}

We adopt the OpenNMT \citep{opennmt} implementations for the first three baselines. The pre-trained Chinese GPT, baselines of \emph{Trans+Pre}, \emph{Trans+Pre+Blo}, \emph{Trans+Pre+Pen} and our models are implemented with Fairseq \citep{fairseq}. We choose Adam~\citep{kingma2014adam} as the optimizer and adopt regular training techniques such as gradient clipping~\citep{clip} and weight decay. The $\beta$ of Adam is adjusted according to different model architectures. During inference, the sampling strategy for all the methods is beam search with a beam size of 10.
\subsection{Evaluation Metrics}
\subsubsection{Automatic metrics.}
\textbf{\textit{Perplexity}} is used to evaluate whether the model assigns a high probability to the unseen reference responses in the validation set. 
\textbf{\textit{F1 score}} is used to evaluate the prediction accuracy of the repetition labels for partner repetition and self-repetition.
\textbf{\textit{Distinct}} \citep{distinct} is used to measure the diversity of the generated responses. Higher \textit{Distinct} score means that the generated response contains more unique \textit{n-gram}s when the length is the same. Here the \textit{Distinct} score of \textit{2-gram} (\textit{Dis-2}) is calculated.   
\textbf{\textit{N-gram overlap}} is the mean number of repeated \textit{n-gram}s between responses and utterances, which is used to evaluate the repetition level. Here \textit{1-gram}, \textit{2-gram}, \textit{3-gram} and \textit{4-gram} are computed, abbreviated as \textit{ov-1}, \textit{ov-2}, \textit{ov-3} and \textit{ov-4}.
\textbf{Beyond \textit{n-gram} Overlap.} \quad \citet{degeneration} presents the phenomenon of repetition feedback loops. Specifically, for any given token, such as \emph{know}, the following sequence of inequality holds:
\begin{equation*}
P(\text{know}|\text{I\,don't}) < P(\text{know}|\text{I\,don't\,know.\,I\,don't})
\end{equation*}
\noindent In the dialogue generation scenario, the first ``I don't know'' can appear in the utterance and the second ``I don't know'' can appear in the response. Once the decoding procedure enters the repetitive pattern, the cycle continues. Obviously, real texts do not repeat in this manner. Therefore, the overlap score of generated responses should decay slower than that of the ground truth. To verify the guess, we define the scalar $r_i$. Given the overlap score \textit{ov-1}, \textit{ov-2}, \textit{ov-3} and \textit{ov-4}, \\
\centerline{$r_1=$\;{\rm \textit{ov-1}}$/${\rm \textit{ov-2}},{$r_2=$\;{\rm \textit{ov-2}}$/${\rm \textit{ov-3}}},{$r_3=$\;{\rm \textit{ov-3}}$/${\rm \textit{ov-4}}}} \\          
$r_i$ for all the baselines are shown in Figure \ref{fig:ratio}. It is clear that the curve of the ground truth locates above those of the generation models except for the plain Transformer; the exception is probably due to Transformer's weak fitting capability that it cannot even capture the repetition pattern. Both the copy mechanism and coverage mechanism aggravate the repetition as expected. Though \emph{Trans+Pre} has the most severe repetition feedback loops, its penalized sampling versions have much lower adjacent overlap scores, which proves that the penalized sampling is beneficial.

We check dozens of generated responses with \textit{4-gram} overlap, and the majority of them are unsatisfactory. The fact demonstrates that \textit{n-gram} repetition of a big $n$ should be avoided. Considering \textit{1-gram} and \textit{2-gram} repetition is reasonable, we believe that if a model has a decreasing trend of overlapping scores from \textit{ov-1} to \textit{ov-4}, then we consider it to be a high quality dialogue generation model.

\subsubsection{Human Metrics.}\quad A series of metrics are proposed in \citep{what} to evaluate the dialogue generation comprehensively. As the dialogue evaluation runs in a single-turn way, we only choose the below metrics.

\emph{Fluency}: Whether the generated response is coherent and human-like.

\emph{Making Sense}: Whether the response is appropriate for the context.

\emph{Interestingness}: Whether the response is boring or unexpected, such as ``I don't know'' is not interesting enough.

\emph{Repetition}: Suitability for repetition given context. Crowdworkers need to judge whether the repetition is appropriate or not.

In order, the rating scale from worst to best is (1, 2, 3, 4). One hundred dialogues are sampled from the test set. The inter-rater annotation agreement between three crowdworkers is measured using the Fleiss's Kappa $\kappa$~\citep{mchugh2012interrater}. The $\kappa$ for \emph{Fluency}, \emph{Making Sense}, \emph{Interetingness} and \emph{Repetition} are 0.32, 0.51, 0.35, 0.72 respectively. The results from different markers are with weak consistency but of similar trends for different models, since there was no consensus between the markers on the crowd-sourcing platform. The Rep metric is more easy to judge and has the best agreement among different workers.

\section{Results and Analysis}
\label{section: Results and Analysis}

\begin{table*}[ht]
  \small
  \centering
  \begin{tabular}{ccccccccccc}
  \toprule[1pt]
  Models  & \textit{ppl} & \textit{Dis-2} & \textit{ov-1}  & \textit{ov-2}  &\textit{ov-3} & \textit{ov-4} & Flu & Mak & Int & Rep \\
  \hline
  Ground Truth  & -    & 0.61   & 1.74  & 0.53   & 0.17   & 0.066 &3.56  & 3.16 & 2.98  & 3.58 \\
  \hline
  Transformer  & 38.56 & 0.052    & 1.48  & 0.48  & 0.15 & 0.057 &3.91  & 2.34  & 1.54 &1.95  \\
  Trans+Cop & 34.26    & 0.37   & 2.28   & 1.25   & 0.60 & 0.29   &3.78  &2.52   &1.61   & 1.53 \\
  Trans+Cop+Cov  & -  & \textbf{0.53}  & 3.26  & 1.93  & 0.98 & 0.48       &3.73  &2.27   &1.56   & 1.47  \\
  \hline
  Trans+Pre       & \textbf{19.53}  & 0.41  &\textbf{3.51}   &\textbf{2.33}  &\textbf{1.51}   &\textbf{0.99}       &3.8   &2.6  &1.75  & 1.81 \\
  Trans+Pre+Blo & -  & 0.19   & 0.83   &0.27  &0.093       &0.041      &3.96  &2.63  &1.73  &2.38 \\
  Trans+Pre+Pen & -  & 0.20   & 0.86 & 0.37   & 0.16       &0.082      &\textbf{3.99}  &2.61  &1.76  &2.2 \\
  \hline
  \bf{ours}    & 19.64   &0.23  & 1.16  &0.32 & 0.034  &0.007    &3.96  &\textbf{2.85}  &\textbf{1.95} &\textbf{2.61}  \\
  \bf{our+blo} & -  & 0.20  & 0.82  & 0.22  & 0.042    &0.014     &3.96  &2.81  &1.78 &2.53 \\
  \bf{our+pen} & - & 0.21   & 0.93  & 0.34  & 0.13     &0.056     &3.94  &2.75  &1.79 &2.25 \\
  \bottomrule[1pt]
  \end{tabular}
  \caption{Automatic evaluation and human evaluation on Chinese Weibo dataset. The automatic metrics include perplexity (\textit{ppl}), mean Length (Len), \textit{Dis-2}, \textit{ov-1}, \textit{ov-2}, \textit{ov-3} and \textit{ov-4}. The human metrics include Fluency (Flu), Making Sense (Mak), Interestingness (Int) and Repetition (Rep).}
  \label{tab:all}
\end{table*}

\noindent \textbf{Effect of Penalized Decoding:}
\begin{table}
  \small
	\centering
	\begin{tabular}{ccccc}
		\toprule 
		\textbf{Method} & \textit{ov-1} & \textit{ov-2} & \textit{ov-4} & \textit{Dis-2} \\ 
		\midrule  
		Trans+Pre+Blo \& \textit{$w_l$, $w_s$}=0.5 &1.52 &0.69 &0.18 &0.25\\
		Trans+Pre+Blo \& \textit{$w_l$, $w_s$}=1.0 &0.54 &0.15  &0.02  &0.16\\
		Trans+Pre+Blo \& \textit{$w_l$, $w_s$}=2.0 &0.07 &5e-3 &5e-4 &0.14\\
		Trans+Pre+Pen \& $\theta_l, \theta_s$=1.1 &1.84 &0.98 &0.32 &0.29\\
		Trans+Pre+Pen \& $\theta_l, \theta_s$=1.3 &0.37 &0.12 &0.02 &0.16\\
		Trans+Pre+Pen \& $\theta_l, \theta_s$=1.4 &0.15 &0.03 &4e-3 &0.15\\
		\bottomrule 
	\end{tabular}
	\caption{The automatic evaluation of \textbf{Trans+Pre+Blo} with different penalizing weight \textit{w} and \textbf{Trans+Pre+Pen} with different penalizing weight $\theta$.}
	\label{tab:block1}
\end{table}
Although Table~\ref{tab:all} show that the penalized decoding sometimes can improve the quality of the generated texts in terms of human evaluation, Table~\ref{tab:block1} shows it is difficult to strike a satisfactory balance between no repeating too much and a high distinct score. When we set the block weight $w$ and penalizing factor $\theta$ to a larger value, the Distinct score and overlap score decrease correspondingly. Similarly, as the repetitions increase, we obtain a higher Distinct score. We manually inspect the responses of low \textit{Dis-2} or high \textit{Dis-2}, finding texts of low \textit{Dis-2} contain more dull responses, which often consist of many high-frequency words. From the above observations, we conclude that texts generated by penalized decoding generally fall into two extremes. We either obtain high distinct texts, but there will be a lot of repetition, or boring low distinct texts. The probability restriction at inference can alleviate excessive repetition but narrow the valid search space. Penalized decoding strikes a trade-off between repetition and diversity. In practice, we find it a sound option to set $w$ to 0.8 and $\theta$ to 1.2. 

\noindent \textbf{Performance of the Classifiers:}
\quad We report the accuracy of our classifiers predicting the repetition label of $T_t$ in Table \ref{tab:acc}, at the same time, validating the importance of the repetition time embedding. Due to the unbalanced nature of the classification task, the classifier's F1 score is relatively low but it still outperforms random guesses significantly, validating the effectiveness of this model. The model of self-repetition and the model of partner repetition have similar F1 scores, indicating that repetition phenomena from different sources may have similar patterns. The repetition time embedding increases the F1 score by a large margin, proving that incorporating the number of repetitions as the additional inputs is truly helpful.

\begin{table}[htbp]{
\centering
  \begin{tabular}{ccc}
        \toprule 
        Classifier & F1  & Random \\ 
        \midrule  
        Self  &0.54 &0.08\\
        + Repetition time &0.59 &0.08 \\
        Partner & 0.52 &0.17\\
        + Repetition time &0.62 &0.17\\
        \bottomrule 
  \end{tabular}
  \caption{ \small The F1 score of predicting repetition labels. \emph{Self} stands for the self-repetition classifier and \emph{Partner} means the partner repetition classifier. \emph{Random} is the baseline of random guessing.
  \label{tab:acc}
  }
  }
  \end{table}

\noindent \textbf{Effect of setting different Repetition Time:}
\quad The number of repeated tokens can be computed accurately at training while it is unavailable at inference. Following the previous explanation, repetition time can be set to a specific value or sampled from the estimated distribution of the training set. Because self-repetition and partner repetition both serve almost the same purpose in our framework, we only test the performance of the partner repetition. The results are illustrated in Table~\ref{tab:inference}. It shows the repetition time embedding controls the repetition degree effectively. Considering sometimes it is necessary to repeat in order to properly respond to the utterance, empirically we find it is sensible to set the repetition time to the average number of repeated tokens. The same law applies to the self-repetition case as well.
\begin{table}[tbh]{
  \centering
  \begin{tabular}{ccccc}
		\toprule 
		Num. Repe. & \textit{ov-1} & \textit{ov-2} & \textit{ov-4} & \textit{Dis-2} \\ 
		\midrule  
		0 &0.0 &0.0 &0.0 &0.14 \\
		1 &0.48 &0.025 &0.0 &0.15\\
		2 &1.16 &0.32  &0.0065  &0.23\\
		3 &2.036 &0.86 &0.085 &0.31\\
		Random &0.68 &0.14 &0.003 &0.17\\
        \bottomrule 
  \end{tabular}
  ~\hfill
  \caption{\small The impact of setting different repetition times for the classifier during inference. \emph{Num. Repe.} denotes the number of repetition time, the input of the repetition time embedding layer. We use \emph{Transformer} as the decoder. The average repetition time on the training dataset is 1.7.
  \label{tab:inference}
  }
  }
  \end{table}

\noindent \textbf{Effect of Classifier Guided Decoding:}
\quad From Table \ref{tab:all}, the valid \textit{ppl} of the \textit{Trans+Pre} and \textbf{ours} are nearly the same, indicating that the newly added classification loss has almost no negative impact on language modeling. Due to the coupling of \textit{Dis-2} and overlap score (Table~\ref{tab:block1}), two models are suitable for comparison only when they have close value on one of the two metrics. Compared to \textit{Trans+Pre+Blo} and \textit{Trans+Pre+Pen}, our model has higher \textit{Dis-2} and \textit{ov-1} values 
but has distinctively lower \textit{ov-3} and \textit{ov-4} values. As previously stated in Section~\textbf{Beyond \textit{n-gram} Overlap} ---  a model is considered decent if it exhibits a decreasing trend of overlapping scores from \textit{ov-1} to \textit{ov-4}. We consider our method better at tackling the repetition issue in dialogue generation. Moreover, our method outperforms all other baselines in human evaluation, though fluency score is not the best but still high enough. Because the classifier is not perfect, we adopt a conservative strategy to tune the thresholds of the classifiers. We ensure the model is prohibited from repeating only when the classifier has high confidence (high recall score for the positive label). We believe these conservative thresholds contribute to producing high-quality texts as well.

\noindent \textbf{Effect of Classifier Enhanced Soft Penalized Sampling:}
\quad From Table \ref{tab:all}, the combination of our model and soft penalized sampling obtains smaller \textit{Dis-2}, \textit{ov-1} and larger \textit{ov-4}. In terms of human evaluation, the performance of our model is also degenerated by this method, though they outperform the baseline methods. We conclude this combination weakens the classifier's ability to control repetition. The biased prior knowledge on which the soft penalized sampling based may contribute to the degenerated performance.

\section{Discussion}
\subsection{Why our Method Works}
The dialogue generation model is autoregressive. During training, it conditions on ground-truth tokens from the previous sequence, while at inference, it conditions on the model-generated sequence. This is known as exposure bias \citep{exposure1,exposure2}. On the other hand, the learned distribution is inaccurate and tokens belonging to tail distribution are merely sampled during inference. One reason of sampling from the top distribution is to guarantee the quality of the generated texts. Consequently, even for the super huge language models, GPT3 \citep{gpt3} and Jurassic-1~\citep{jurassic}, only tokens from the top distribution are sampled. This is another gap between training and inference. For a dialogue generation model, beam search (our adopted decoding method), smaller $k$ in top-k, or smaller $p$ in top-p are choices to enhance the semantic relevance between utterance and response, but they also lead to excessive repetition. The problem disappears, i.e., the repetition metric of the model generated texts matches that of gold text, only when generating from untruncated distribution~\citep{massive}. Otherwise, the sampling distribution during inference, e.g. for top-p sampling, is distorted compared to the distribution learned from the training data. Therefore, it is beneficial to introduce a non-distorted signal to assist the distorted decoding strategy. In light of this, we train a classifier to directly learn the repetition distribution in text, which is employed to control the repetition when sampling a word. The data distribution learned by the classifier during training is consistent with the input distribution during inference.

\subsection{Repetition and Diversity}
Compared to repetition, the metric of diversity receives more attention \citep{diversity1,diversity2,diversity3,what,diversity4}. With the advent of dialogue generation models based on pre-training, these two metrics are often coupled in experimental results, thus they can be boiled down to a single one~\citep{massive}. However, users' perception of these two metrics is different~\citep{metric}. Inappropriate repetition in the text is clearly perceptible while the lack of diversity is more discreet and not easily detectable. A large $k$ in top-k sampling entails more repetition and diversity simultaneously but hurts the semantic rationality~\citep{massive}. Such sampling strategy strikes a trade-off between repetition, diversity, and semantic relevance. In contrast, when adopting top-k sampling of a smaller $k$, our method allows a model to generate relatively diverse and semantic-relevant responses with reasonable repetition.

When the pretraining dialogue model scales to a larger size, it supports a more adventurous sampling strategy. For instance, Meena~\citep{meena} and BlenderBot~\citep{blenderbot} both adopt top-p sampling while weaker dialogue models are more suitable with beam search. The responses of these giant models are naturally less repetitive and more diverse. However, \citet{gpt3} still report semantic repetition at the document level and we encounter severe word repetitions in the demo of Jurassic-1 \citep{jurassic}. Based on the above evidence, huge language models may need repetition control as well.    

\section{Conclusion}
Some generated responses from the state-of-the-art dialogue generation models are still suffering from excessive repetitions. Penalized sampling is an effective remedy but it relies on setting a proper static weight. Setting it too high makes the model more inclined to generate tokens unaware of the context while setting it too low is inherently contrary to the goal of reducing repetition and causes repetition feedback loop. We study the limitation of penalized sampling and design a classifier, using its output as a non-distorted signal to guide repetition during inference. The classifier leverages both the context and the history of repetition statistics; we use it to assist the dialogue generation model by explicitly determining when to conduct penalized sampling during decoding. Experimental results on the single-turn Chinese dialogue dataset show our method outperforms all the strong baselines on both automatic and human evaluation. Particularly, our method is effective in reducing the repetition problem caused by mainstream decoding methods while persevering the diversity of texts at the same time.

We appeal to explore more issues at inference following the paradigm in this paper, for example, investigating the trade-offs of semantic reasonableness and diversity. In addition, the effectiveness of our framework needs to be validated on general text generation tasks as well.

\bibliography{ref}
\end{document}